%% file: acl_latex.tex
\documentclass[11pt]{article}

\PassOptionsToPackage{table}{xcolor}

\usepackage[preprint]{acl}

\usepackage{times}
\usepackage{latexsym}
\usepackage[T1]{fontenc}
\usepackage[utf8]{inputenc}
\usepackage{microtype}
\usepackage{inconsolata}
\usepackage{graphicx}

\usepackage{amsmath}
\usepackage{amssymb}
\usepackage{booktabs}
\usepackage{multirow}
\usepackage{pifont}
\usepackage{enumitem}
\usepackage{xspace}
\usepackage[most]{tcolorbox}

\definecolor{seqblue}{HTML}{6DA7EC}
\definecolor{boxframe}{HTML}{9EC5F4}
\definecolor{boxhead}{HTML}{CDE2FB}
\definecolor{good}{HTML}{0CA30C}
\definecolor{critical}{HTML}{D03B3B}
\definecolor{inkmuted}{HTML}{52514E}

\newcommand{\hc}[2]{\cellcolor{seqblue!#1}#2}
\newcommand{\cmark}{\textcolor{good}{\ding{51}}}
\newcommand{\xmark}{\textcolor{critical}{\ding{55}}}
\newcommand{\qmark}{\textcolor{inkmuted}{\textbf{?}}}
\newcommand{\dataset}{\textsc{HomogeneityNLI}\xspace}
\newcommand{\sem}[1]{[\![\text{#1}]\!]}

\newtcolorbox{groupbox}[1]{enhanced, colback=white, colframe=boxframe, colbacktitle=boxhead,
  coltitle=black, fonttitle=\bfseries\small, title={#1}, boxrule=0.6pt, arc=1.5mm,
  left=4pt, right=4pt, top=2pt, bottom=2pt, fontupper=\small, before skip=5pt, after skip=5pt}
\newtcolorbox{promptbox}[1]{enhanced, colback=white, colframe=boxframe, colbacktitle=boxhead,
  coltitle=black, fonttitle=\bfseries\small, title={#1}, boxrule=0.6pt, arc=1.5mm,
  left=4pt, right=4pt, top=3pt, bottom=3pt, fontupper=\small, before skip=6pt, after skip=6pt}

\newcommand{\projecturl}{https://duaanakshbandi.github.io/homogeneity-nli/}
\newcommand{\dataurl}{https://huggingface.co/datasets/duaanakshbandi/HomogeneityNLI}
\makeatletter
\newcommand{\ifanonymous}[2]{\ifacl@anonymize #1\else #2\fi}
\makeatother

\title{Neither True nor False: \\Do LLMs Represent Truth-Value Gaps in Plural Predication?}

\author{
  Duaa Nakshbandi$^1$ \quad Zhiyu An$^1$ \quad Yiping Li$^1$ \quad Sachin Giri$^1$ \quad Wan Du$^1$ \\
  $^1$University of California, Merced \\
  \texttt{\{duaanakshbandi,zan7,yipingli,sachingiri,wdu3\}@ucmerced.edu}
}

\begin{document}
\maketitle
\begin{abstract}
Large Language Models (LLMs) are increasingly being used to summarize and verify claims in critical sources that inform consequential decisions.
Such claims are full of plural definites.
If \textit{three of five patients} improved after taking drug A, the claim \textit{the patients improved after drug A} is not true, since two did not, yet its negation, \textit{the patients didn't improve after drug A}, is not true either, since three did.
In formal semantics, this truth-value gap is known as the \textbf{Homogeneity Gap}: \emph{both} the plural definite and its negation are neither true nor false when the property holds for only a fraction of the plural referent.
A model that cannot represent this gap turns mixed evidence into a clean verdict, an error that can be costly in high-stakes use.
We introduce \dataset, a diagnostic dataset of 1,000 minimal-pair items crossing determiner, polarity, and group composition across ten logical conditions.
Evaluating seven open-weight models, we uncover a pervasive \textbf{Bivalence Bias}: models assign plural definites their homogeneous truth conditions.
In the zero-shot setting, no model recognizes the gap for both polarities in more than 15\% of scenarios, and negated definites are judged \textit{False} in 84\% of cases.
Explicit linguistic rules, chain-of-thought, and enumeration prompts produce unstable, model-specific effects that often entrench the bias.
Our findings indicate that current open-weight LLMs acquire the truth conditions of plural predication but not its trivalent logic, and cannot yet be trusted to summarize or verify claims whose evidence is mixed.\footnote{\ifanonymous{Data and code will be made publicly available upon publication.}{Project page: \url{\projecturl}. Data: \url{\dataurl}.}}
\end{abstract}

\section{Introduction}

\begin{table}[t]
  \centering
  \small
  \setlength{\tabcolsep}{10pt}
  \begin{tabular}{@{}l p{3.5cm} c@{}}
    \toprule
    \multicolumn{3}{@{}l}{\textbf{P:} Five patients took drug A. Three of them improved} \\
    \multicolumn{3}{@{}l}{\phantom{\textbf{P:}}  after taking drug A and two did not.} \\
    \midrule
    \textbf{Determiner} & \textbf{Hypothesis} & \textbf{Label} \\
    \midrule
    \textbf{the} & \textbf{H:} The patients \textit{improved} \newline after taking drug A. & \qmark\ Neither \\
    \textbf{the} + not & \textbf{H:} The patients \textit{didn't \newline improve} after taking drug A. & \qmark\ Neither \\
    \textbf{all} & \textbf{H:} \textit{All} the patients improved after taking drug A. & \xmark\ False \\
    \textbf{not all} & \textbf{H:} \textit{Not all} the patients improved after taking drug A. & \cmark\ True \\
    \bottomrule
  \end{tabular}
  \caption{Examples of the Homogeneity Gap. Given a mixed group, a plural definite and its negation are neither true nor false, while their quantified counterparts are classically false or true.}
  \label{tab:intro}
\end{table}

Large Language Models (LLMs) have demonstrated remarkable proficiency in natural language inference (NLI) and complex reasoning tasks \citep{brown-etal-2020-language,touvron-etal-2023-llama}.
Despite these advancements, questions remain regarding the robustness of their reasoning: do models genuinely grasp compositional semantics, or do they rely on surface-level probabilistic heuristics \citep{bender-etal-2021-dangers,dziri-etal-2023-faith}?
This question is pressing where the logic of natural language departs from textbook logic.
A faithful reasoning agent must know not only \emph{when} a sentence is true or false, but also \emph{when it is neither}.

Consider the sentence ``\textit{The patients improved after taking drug A}'' in a situation where three of five patients improved after taking drug A (Table~\ref{tab:intro}).
The sentence is not true: two patients did not improve after taking drug A.
Yet its negation, ``\textit{The patients didn't improve after taking drug A}'', is not true either: three patients did.
Plural definite descriptions thus give rise to a \emph{truth-value gap} whenever the group is not uniform with respect to the predicate, a phenomenon known in formal semantics as \textbf{Homogeneity} \citep{lobner-1987-conceptual,schwarzschild-1994-plurals,kriz2015aspects}.
In contrast, overtly quantified sentences leave no gap: ``\textit{All the patients improved after taking drug A}'' is simply false, and ``\textit{Not all the patients improved after taking drug A}'' is simply true.
Adults often judge both definite sentences in such situations as neither completely true nor completely false \citep{kriz-chemla-2015-two}, while some preschoolers first interpret \textit{the patients} as \textit{some patients} \citep{tieu-etal-2015-acquisition}.
A robust reasoner must therefore respect a three-way boundary: accept a plural predication when the whole group has the property, reject it when no member does, and suspend judgment in between.

We hypothesize that LLMs instead exhibit a \textbf{Bivalence Bias}: a tendency to force every sentence into \textit{True} or \textit{False}, collapsing the homogeneity gap into falsity.
This bias may stem from the supervision conventions of NLI, where a hypothesis either follows from, contradicts, or is underdetermined by its premise \citep{bowman-etal-2015-large}; none of these categories fits a sentence whose premise fully specifies the facts but whose truth value is nonetheless undetermined.
Consequently, models may conflate \emph{not true} with \emph{false}.
Prior work has probed inference models on negation \citep{kassner-schutze-2020-negated,hossain-etal-2020-analysis}, implicature and presupposition \citep{jeretic-etal-2020-natural}, and syntactic heuristics \citep{mccoy-etal-2019-right}, but it remains unexplored whether generative models can represent a truth-value gap at all.

To bridge this gap, we introduce \dataset, a diagnostic dataset to probe homogeneity.
We construct minimal pairs across ten logical conditions, crossing the determiner (\textit{the}, \textit{all}, \textit{some}), polarity, and group composition.
We evaluate state-of-the-art open-weight models and uncover the following critical insights:

\begin{enumerate}[label=\arabic*),leftmargin=*,itemsep=2pt plus 0pt minus 0pt,parsep=0pt,topsep=3pt plus 0pt minus 0pt,partopsep=0pt]
  \item \textbf{Bivalence Bias Dominates Model Inference:} Models almost never judge both a plural definite and its negation as neither true nor false. Crucially, they are near-perfect on quantified and uniform-context controls, revealing a failure of \emph{truth-value assignment} rather than of comprehension.
  \item \textbf{Prompting Interventions Yield Unstable Effects:} Explicit rules, chain-of-thought, and enumeration of group members shift models between collapse and partial gap recognition in model-specific ways, and often entrench the collapse.
  \item \textbf{Truth Conditions and Truth Values Are Dissociated:} Model rationales restate the positive definite as \textit{all} and its negation as \textit{none}: models compute homogeneous truth conditions and then evaluate them bivalently, while showing a human-like tolerance for exceptions in nearly uniform groups.
\end{enumerate}

\section{Background}
\label{sec:background}

\subsection{Plural Predication and the Homogeneity Gap}

Plural definite descriptions (\textit{the patients}) refer to a group as a whole.
With a \emph{distributive} predicate, which applies to each member individually (e.g., \textit{improved}, \textit{be red}), the truth of the sentence depends on how the property is distributed across the group \citep{schwarzschild-1994-plurals}.
Quantified descriptions (\textit{all the patients}, \textit{some of the patients}) have classical truth conditions, so that a sentence and its negation are complementary:
\begin{equation*}
  \sem{not all the $X$ $P$} = 1 - \sem{all the $X$ $P$}
\end{equation*}
Plural definites, in contrast, have \emph{strong} truth conditions in both polarities \citep{lobner-2000-polarity,gajewski-2005-neg}.
For a plurality $X$ and a distributive predicate $P$:
\begin{equation}
  \label{eq:homogeneity}
  \sem{the $X$ $P$} =
  \begin{cases}
    1 & \text{if } \forall x \in X: P(x)\\
    0 & \text{if } \forall x \in X: \neg P(x)\\
    \# & \text{otherwise}
  \end{cases}
\end{equation}
and symmetrically for \textit{the $X$ not-$P$}, where $\#$ marks a truth-value gap.
In a \emph{mixed} situation, complementarity therefore fails:
\begin{equation*}
  \sem{the $X$ not-$P$} \neq 1 - \sem{the $X$ $P$}
\end{equation*}
Both definite sentences are neither true nor false, while \textit{all the $X$ $P$} is false and \textit{not all the $X$ $P$} is true.

The gap has been derived as a homogeneity presupposition \citep{schwarzschild-1994-plurals,lobner-2000-polarity}, as supervaluation over an existential and a universal reading \citep{spector-2013-homogeneity,kriz2015aspects}, and as recursive strengthening of a literal existential meaning by implicature \citep{magri-2014-account}.
All accounts acknowledge that homogeneity is not absolute: in suitable contexts, \textit{the patients improved} is accepted despite a few exceptions, a phenomenon known as \emph{non-maximality} \citep{brisson-1998-distributivity,lasersohn-1999-pragmatic,malamud-2012-meaning,kriz-2016-homogeneity}.
Experimentally, \citet{tieu-etal-2015-acquisition} identified three interpretive profiles from the answers to positive and negated definites in mixed situations: \emph{homogeneous} (both rejected), \emph{existential} (\textit{the} $=$ \textit{some}: positive accepted, negated rejected), and \emph{universal} (\textit{the} $=$ \textit{all}, with negation over it: positive rejected, negated accepted).
Adults were mostly homogeneous, while a subgroup of 4--5-year-olds was existential.

\subsection{Related Work: Quantification, Negation, and Truth-Value Gaps in NLP}

Evaluation of semantic competence has evolved from large NLI benchmarks \citep{bowman-etal-2015-large} to targeted minimal-pair diagnostics \citep{warstadt-etal-2020-blimp}.
Such diagnostics revealed reliance on lexical-overlap heuristics \citep{mccoy-etal-2019-right} and persistent difficulties with negation \citep{ettinger-2020-bert,kassner-schutze-2020-negated}, which is also underrepresented in NLI benchmarks \citep{hossain-etal-2020-analysis}.
Closer to non-classical meaning, \citet{jeretic-etal-2020-natural} probed scalar implicatures and presuppositions in NLI models, and \citet{pavlick-kwiatkowski-2019-inherent} showed that humans systematically disagree on inference labels. To our knowledge, no prior study has systematically evaluated whether LLMs respect the Homogeneity Gap.

\section{Dataset - \texorpdfstring{\dataset}{HomogeneityNLI}}
\label{sec:dataset}

To evaluate LLMs' ability to reason about homogeneity and truth-value gaps, we introduce \dataset, a diagnostic NLI dataset in English.

\paragraph{Dataset Construction with Different Predicates.}
The construction of \dataset employs a template-based approach to ensure syntactic uniformity while isolating semantic reasoning.
We curated 100 scenarios, each pairing a plural noun phrase with a distributive predicate, balanced across four predicate classes (25 each): \textbf{Action} (e.g., \textit{laughed, passed the exam}), \textbf{Property} (e.g., \textit{were red, were made of glass}), \textbf{Location} (e.g., \textit{were in the garage}), and \textbf{State} (e.g., \textit{were asleep, were frozen}).
We excluded collective predicates (e.g., \textit{gathered}), which interact with homogeneity differently \citep{kriz2015aspects}.
Each scenario introduces a group of $N \in \{4, 5, 6, 8, 10\}$ members (``\textit{Five hearts were drawn on the card.}'') and specifies how many of them, $k$, have the property in mixed contexts.
To probe non-maximality,we draw $k$ from one of three proportion bins, \emph{1 of N}, \emph{about half} ($\lceil N/2 \rceil$), and \emph{N$-$1 of N}, balanced across predicate classes (full lexicon in Appendix~\ref{app:lexicon}).
From each scenario we derive ten items that share the same group and predicate and differ only in the context sentence and the hypothesis, resulting in exact minimal pairs and a total of 1,000 examples.
Uniform contexts use ``\textit{Every one of them}'' rather than \textit{all}, to avoid lexical overlap with quantified hypotheses \citep{mccoy-etal-2019-right}.

\begin{table*}[t]
  \centering
  \small
  \setlength{\tabcolsep}{5pt}
  \begin{tabular}{@{}c l c l c@{}}
    \toprule
    \textbf{Group} & \textbf{Condition} & \textbf{Context} & \textbf{Hypothesis} & \textbf{Gold Label} \\
    \midrule
    \textbf{A} & \textbf{Gap Definite} (\textit{the} + P, mixed group) & GAP & The hearts were red. & Neither (\qmark) \\
    \textbf{B} & \textbf{Gap Negated Definite} (\textit{the} + not-P, mixed group) & GAP & The hearts weren't red. & Neither (\qmark) \\
    \textbf{C} & \textbf{Gap Universal} (\textit{all} + P, mixed group) & GAP & All the hearts were red. & False (\xmark) \\
    \textbf{D} & \textbf{Gap Negated Universal} (\textit{not all} + P, mixed group) & GAP & Not all the hearts were red. & True (\cmark) \\
    \midrule
    \textbf{E} & \textbf{Uniform Definite} (\textit{the} + P, all members) & ALL & The hearts were red. & True (\cmark) \\
    \textbf{F} & \textbf{Uniform Negated Definite} (\textit{the} + not-P, all members) & ALL & The hearts weren't red. & False (\xmark) \\
    \textbf{G} & \textbf{Empty Definite} (\textit{the} + P, no member) & NONE & The hearts were red. & False (\xmark) \\
    \textbf{H} & \textbf{Empty Negated Definite} (\textit{the} + not-P, no member) & NONE & The hearts weren't red. & True (\cmark) \\
    \midrule
    \textbf{I} & \textbf{Implicature Probe} (\textit{some} + P, all members) & ALL & Some of the hearts were red. & True$^{*}$ (\cmark) \\
    \textbf{J} & \textbf{Existential Control} (\textit{some} + P, mixed group) & GAP & Some of the hearts were red. & True (\cmark) \\
    \bottomrule
  \end{tabular}
  \caption{The ten logical conditions of \dataset, illustrated with one scenario. All premises begin with ``\textit{Five hearts were drawn on the card.}'', followed by ``\textit{Three of them were red and two were not.}'' (GAP), ``\textit{Every one of them was red.}'' (ALL), or ``\textit{None of them were red.}'' (NONE). Groups A and B are the critical probe, where homogeneity implies a truth-value gap, but a bivalent reasoner is forced to choose \textit{True} or \textit{False}. $^{*}$\textit{True} under the literal meaning of \textit{some}; \textit{False} under the scalar implicature \textit{some but not all}.}
  \label{tab:dataset}
\end{table*}

\paragraph{Homogeneity Logical Conditions.}
The core contribution of \dataset is its minimal-pair design, which crosses the \textbf{Determiner} of the hypothesis (\textit{the}, \textit{all}, \textit{some}) and its \textbf{Polarity} with the \textbf{Context}: a mixed group (GAP), or a group in which all (ALL) or none (NONE) of the members have the property.
The resulting groups A--J are illustrated in Table~\ref{tab:dataset}.

\begin{groupbox}{Groups A \& B: Gap Definites (Target).}
The primary testbed. The premise describes a mixed group; the hypothesis is a positive (A) or negated (B) plural definite. Under homogeneity, both are \textbf{Neither} true nor false. An existential reading (\textit{the} $=$ \textit{some}) makes A \textit{True}; a universal reading with negation over it makes B \textit{True}.
\end{groupbox}

\begin{groupbox}{Groups C \& D: Gap Quantifiers (Baseline).}
Same premise with an overt universal: \textit{All the hearts were red} (C) is \textbf{False}; \textit{Not all the hearts were red} (D) is \textbf{True}. D differs from B only in the scope of negation, isolating homogeneity from negation processing.
\end{groupbox}

\begin{groupbox}{Groups E--H and I \& J: Controls.}
Uniform contexts (E--H) test plural definites when no gap arises, paralleling the controls of \citet{tieu-etal-2015-acquisition}. \textit{Some of the hearts were red} is \textbf{True} in mixed contexts (J) and literally \textbf{True} in ALL contexts (I), where it is \textit{False} under the scalar implicature \textit{some but not all} \citep{noveck-2001-children}, linked to homogeneity by \citet{magri-2014-account}.
\end{groupbox}

Instead of the NLI labels \textit{Entailment}, \textit{Contradiction}, and \textit{Neutral} \citep{bowman-etal-2015-large}, we use the natural-language truth values \textbf{``True''}, \textbf{``False''}, and \textbf{``Neither''}.
We deliberately avoid \textit{Unknown}: in gap contexts the premise fully specifies the facts, so what is missing is not information but a classical truth value, in line with the ternary response options of \citet{kriz-chemla-2015-two}.

\paragraph{Quality Assurance.}
Items are constructed from hand-written templates with explicit agreement handling (Appendix~\ref{app:construction}), and all items were inspected for grammaticality and naturalness.

\section{Experimental Settings}

\subsection{Models}
We evaluate the seven instruction-tuned open-weight models (7--9B parameters) 
\texttt{Llama-3.1-8B-Instruct} \citep{grattafiori-etal-2024-llama},
\texttt{Mistral-7B-Instruct-v0.3} \citep{jiang2023mistral},
\texttt{Qwen2.5-7B-Instruct} \citep{qwen-2025-qwen25},
\texttt{deepseek-llm-7b-chat} \citep{deepseek-2024-llm},
\texttt{gemma-2-9b-it} \citep{gemma-2024-gemma2},
\texttt{glm-4-9b-chat-hf} \citep{glm-2024-chatglm}, and
\texttt{Yi-1.5-9B-Chat} \citep{young2024yi}.

\subsection{Prompts}
To disentangle intrinsic biases from instruction following, we adapt the four prompting strategies.
\textbf{(1) Strict Logic Prompt (Zero-Shot)} frames the model as a ``strict logician'' that outputs \textit{True}, \textit{False}, or \textit{Neither} (``the Premise makes the Hypothesis neither clearly true nor clearly false''), using only the provided text.
\textbf{(2) Definition-Aware Prompt (DAP)} additionally states the homogeneity rule for plural definites and its absence for \textit{all} and \textit{not all}, testing whether failures stem from missing \emph{knowledge} or failed \emph{application}.
\textbf{(3) Chain-of-Thought Prompt (CoT)} \citep{wei-etal-2022-chain} asks the model to analyze how many members have the property (all, some, or none) and what the hypothesis claims before labeling.
\textbf{(4) Enumeration Prompt} draws on the supervaluationist view that plural predication is evaluated against individual members \citep{spector-2013-homogeneity,kriz2015aspects}: the model first lists each member and whether the property holds for it. This is the most direct intervention.

\subsection{Metrics}

\paragraph{(1) Group-wise Accuracy (ACC):}
Classification accuracy for each group A--J; \textbf{Ctrl} is the mean accuracy on the uniform-context controls E--H.

\paragraph{(2) Existential and Universal Bias Rates (EBR, UBR):}
Since the gold label of groups A and B is \textit{Neither}, \textit{True} for the positive definite reveals an existential reading, and \textit{True} for the negated definite a universal reading \citep{tieu-etal-2015-acquisition}: $\text{EBR} = P(\hat{y} = \text{True} \mid A)$ and $\text{UBR} = P(\hat{y} = \text{True} \mid B)$, where $\hat{y}$ is the model's prediction.

\paragraph{(3) Homogeneity Gap Rate (HGR) and Bivalent Collapse Rate (BCR):}
The pair of answers to the positive and negated definite of a scenario $s \in S$ reveals how a model treats homogeneity.
We classify each pair $(\hat{y}^s_A, \hat{y}^s_B)$ into six patterns: \emph{Gap} (Neither, Neither), \emph{Mixed} (one Neither, one False), \emph{Collapse} (False, False), \emph{Existential} (True, not True), \emph{Universal} (not True, True), and \emph{Inconsistent} (True, True).
The first three jointly form the homogeneous profile of a binary judgment task, in which both sentences are rejected \citep{tieu-etal-2015-acquisition}; our three-way labels allow us to split it:
\begin{equation}
  \begin{aligned}
    \text{HGR} &= \tfrac{1}{|S|} \textstyle\sum_{s \in S} \mathbb{I}(\hat{y}^s_A = \hat{y}^s_B = \text{Neither})\\
    \text{BCR} &= \tfrac{1}{|S|} \textstyle\sum_{s \in S} \mathbb{I}(\hat{y}^s_A = \hat{y}^s_B = \text{False})
  \end{aligned}
\end{equation}
A high HGR indicates genuine recognition of the gap; a high BCR indicates that the model computes the truth conditions of Eq.~(\ref{eq:homogeneity}) but assigns falsity whenever they fail, the signature of the Bivalence Bias.

\paragraph{(4) Negation Asymmetry Gap ($\Delta_{NA}$):}
$\Delta_{NA} = \text{ACC}_D - \text{UBR}$ measures whether the model distinguishes \textit{not all the patients improved after taking drug A} (true) from \textit{the patients didn't improve after taking drug A} (not true) in the same situation, the contrast that defines homogeneity; a low value suggests that it treats \textit{the} as \textit{all} under negation.

\paragraph{(5) Scalar Implicature Rate (SI):}
The share of group I items not judged \textit{True}, i.e., the rate at which \textit{some} is read as \textit{some but not all} \citep{noveck-2001-children,fox-2007-free}.

\section{Results and Analysis}

We aim to answer four key questions: whether LLMs recognize the homogeneity gap, how prompting affects this ability, how the gap interacts with group composition and predicate semantics, and what model rationales reveal about the source of failure.

\input{tables/main_results}

\subsection{The Dominance of Bivalence Bias: \textit{``If Not All, Then False''} in the  Zero-Shot Setting.}

\begin{figure}[t]
  \centering
  \includegraphics[width=\columnwidth]{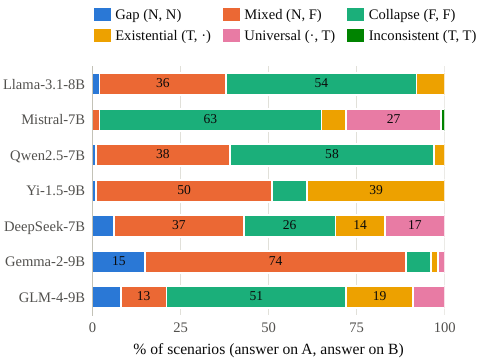}
  \caption{Response patterns on the target pair (answer on A, answer on B) in the zero-shot setting, in \% of the 100 scenarios per model. N = \textit{Neither}, F = \textit{False}, T = \textit{True}. Only Gap reflects the truth-value gap; Collapse and Mixed would both count as ``homogeneous'' rejections in a binary judgment task.}
  \label{fig:patterns}
\end{figure}

The first rows of Table~\ref{tab:main} and Figure~\ref{fig:patterns} present the zero-shot results.
They reveal a fundamental inability of current LLMs to recognize that a plural predication can be neither true nor false.

\paragraph{\texorpdfstring{$P(\text{Gap} \mid \text{Mixed Group}) \approx 0$}{P(Gap | Mixed Group) = 0} ? -- The Illusion of Competence.}
At first glance, models appear competent with plural descriptions, achieving near-perfect accuracy on the quantifier baselines (Acc C $\geq 0.96$, Acc D $\geq 0.99$) and the uniform-context controls (Ctrl $\geq 0.95$), and predicting \textit{Neither} for at most 1.4\% of the items in any of the groups C--J.
However, this competence does not extend to the target pair.
The Homogeneity Gap Rate ranges from 0.00 (\texttt{Mistral}) to 0.15 (\texttt{gemma-2}), with a mean of 0.05: in 95\% of scenarios, models fail to treat both definites as neither true nor false.
The high $\Delta_{NA}$ ($\geq 0.72$) shows that models do distinguish \textit{the patients didn't improve after taking drug A} from \textit{not all the patients improve after taking drug A}; what they miss is the third truth value.

\paragraph{Collapse into Falsity.}
For four of seven models, the most frequent pattern is to judge \emph{both} sentences \textit{False} (BCR: \texttt{Mistral} 0.63, \texttt{Qwen2.5} 0.58, \texttt{Llama-3.1} 0.54, \texttt{GLM-4} 0.51).
Such a model reads \textit{the patients improved after taking drug A} as \textit{all the patients improved after taking drug A} and \textit{the patients didn't improve after taking drug A} as \textit{none of the patients improved after taking drug A}, computing the homogeneous truth conditions of Eq.~(\ref{eq:homogeneity}), but assigns \textit{False} whenever they are not met.
In a binary judgment task \citep{crain-thornton-1998-investigations}, this would pass as homogeneity: 61--97\% of scenarios fall into the ``reject both'' category (Appendix~\ref{app:results}), comparable to the 23 of 25 adults classified as homogeneous by \citet{tieu-etal-2015-acquisition}, yet almost none of them involves an actual gap.

\paragraph{Polarity Asymmetry.}
When models use \textit{Neither}, they reserve it mostly for the positive sentence.
Pooled over models, the positive definite receives \textit{Neither} in 40\% of cases, whereas the negated definite receives it in only 8\% and is judged \textit{False} in 84\% (McNemar's test: 246 vs.\ 24 discordant pairs, $p < 10^{-40}$).\footnote{A two-way repeated-measures ANOVA over the seven models, with polarity and proportion bin (\S\ref{sec:finegrained}) as within-model factors, confirms a main effect of polarity on the rate of \textit{Neither} ($F(1,6) = 14.71$, $p < 0.01$).}
The resulting Mixed pattern (\textit{Neither}, \textit{False}) is the most frequent one for \texttt{gemma-2} (68\% of scenarios) and \texttt{Yi-1.5} (50\%).
This suggests that the negated definite is evaluated as a universal negative claim, which a single counterexample falsifies, whereas the positive sentence invites considerations of partial truth, echoing known difficulties of language models with negation \citep{ettinger-2020-bert,kassner-schutze-2020-negated}.

\paragraph{Model-Specific Readings Mirror Developmental Profiles.}
\texttt{Yi-1.5} accepts the positive definite in 39\% of scenarios while rejecting the negated one (EBR $= 0.39$), the \emph{existential} profile of preschoolers who read \textit{the patients} as \textit{some patients} \citep{tieu-etal-2015-acquisition}.
\texttt{Mistral} shows the mirror image (UBR $= 0.28$), the \emph{universal} profile that \citet{tieu-etal-2015-acquisition} observed in some adults.
No model, however, exhibits the adult-like gap as its dominant pattern.

\paragraph{The Outlier: \texttt{gemma-2} Approaches the Adult Pattern.}
\texttt{gemma-2} shows the highest HGR (0.15), the lowest BCR (0.07), and an accuracy of 0.84 on the positive definite.
Yet even \texttt{gemma-2} judges the negated definite \textit{False} in most scenarios (Acc B $= 0.21$): the polarity asymmetry persists in the most gap-sensitive model.

\subsection{Prompting: Unstable Effects of Rules, Reasoning, and Enumeration}
\label{sec:prompting}

\begin{figure*}[t]
  \centering
  \includegraphics[width=\textwidth]{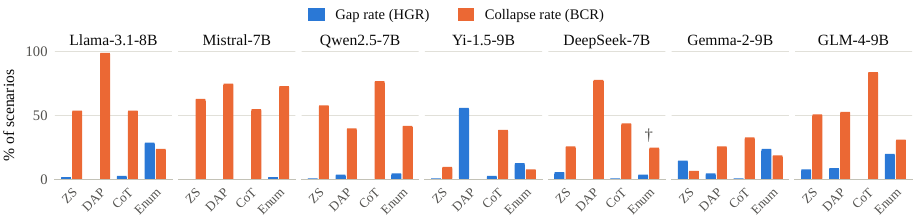}
  \caption{Homogeneity Gap Rate (HGR: both sentences \textit{Neither}) and Bivalent Collapse Rate (BCR: both \textit{False}) per model across prompting strategies (ZS: Zero-shot, Enum: Enumeration). A genuine gap exceeds half of the scenarios only for \texttt{Yi-1.5} with DAP. $^\dagger$Degenerate run in which the uniform-context controls fail (\S\ref{sec:prompting}).}
  \label{fig:prompts}
\end{figure*}

Advanced prompting strategies reveal no consistent route to the gap.
Averaged over models, the HGR moves from 0.05 (zero-shot) to 0.11 (DAP), 0.01 (CoT), and 0.14 (Enumeration), while the BCR rises from 0.38 to 0.53 under DAP and 0.55 under CoT (Figure~\ref{fig:prompts}).

\paragraph{DAP: The Knowledge--Application Gap.}
Supplying the homogeneity rule yields the largest improvement in our study, but for one model only: \texttt{Yi-1.5} judges both sentences \textit{Neither} in 56\% of scenarios (from 1\%) and no longer collapses (BCR $0.10 \rightarrow 0.00$).
For the other models, the rule is ineffective or counterproductive: \texttt{Llama-3.1} collapses almost completely (BCR $0.54 \rightarrow 0.99$; Acc A $0.37 \rightarrow 0.01$), as do \texttt{DeepSeek} ($0.26 \rightarrow 0.78$) and \texttt{Mistral} ($0.63 \rightarrow 0.75$), while \texttt{gemma-2} loses most of its gap judgments (HGR $0.15 \rightarrow 0.05$).
Told that both sentences are ``neither true nor false'', these models appear to retain only that neither is \emph{true}.
The persistence of the bias suggests that bivalence is a robust feature of the models' output distribution that in-context rules do not override \citep{dziri-etal-2023-faith}.

\paragraph{CoT: Reasoning Entrenches Bivalence.}
Step-by-step reasoning suppresses the gap for every model (HGR $\leq 0.03$) and increases the collapse for five of seven models (e.g., \texttt{GLM-4} $0.51 \rightarrow 0.84$, \texttt{Qwen2.5} $0.58 \rightarrow 0.77$, \texttt{gemma-2} $0.07 \rightarrow 0.33$).
When analyzing the group, models restate the hypothesis with an explicit quantifier, converting a homogeneous predication into a bivalent universal claim.\footnote{E.g., for \textit{The dogs barked} with five of ten dogs barking, \texttt{Mistral} reasons: ``\textit{The hypothesis claims that all dogs barked. Since not all dogs in the group barked, the hypothesis is false.}''}
CoT also introduces sporadic scalar implicatures (SI $= 0.15$ for \texttt{Llama-3.1}, $0.14$ for \texttt{DeepSeek}) and lowers \texttt{Llama-3.1}'s accuracy on the \textit{all} baseline (Acc C $= 0.74$).

\paragraph{Enumeration: A Partial Release at a Cost.}
Listing the members before judging yields the highest average HGR, with gains for \texttt{Llama-3.1} ($0.02 \rightarrow 0.29$), \texttt{gemma-2} ($0.15 \rightarrow 0.24$), \texttt{GLM-4} ($0.08 \rightarrow 0.20$), and \texttt{Yi-1.5} ($0.01 \rightarrow 0.13$).
However, it destabilizes the controls: \texttt{Llama-3.1}'s Ctrl drops from 1.00 to 0.79, and \texttt{DeepSeek} degenerates entirely (Ctrl $= 0.62$, Acc D $= 0.11$; all definite controls correct in only 10 of 100 scenarios), so we exclude this run from pooled analyses.

\paragraph{Summary of Prompting Results.}
Current LLMs cannot be prompted into a trivalent logic: prompts move models between \textbf{collapse} (both \textit{False}) and \textbf{partial gaps} (\textit{Neither} for the positive sentence only), and a gap for both polarities reaches at least 20\% of scenarios in only four of 28 model--prompt combinations.

\subsection{Fine-grained Analysis on Group Composition and Predicates}
\label{sec:finegrained}

We disaggregate the zero-shot answers to the target pair by the proportion of members with the property and by predicate class (Figures~\ref{fig:ratio} and~\ref{fig:class}; per-model results in Appendix~\ref{app:results}).

\begin{figure*}[t]
  \centering
  \begin{minipage}[t]{0.48\textwidth}
    \centering
    \includegraphics[width=\linewidth]{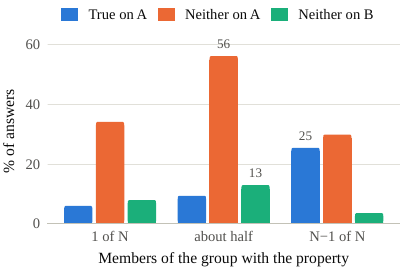}
    \caption{Answers to the target pair by the proportion of members with the property (zero-shot, pooled over models).}
    \label{fig:ratio}
  \end{minipage}\hfill
  \begin{minipage}[t]{0.48\textwidth}
    \centering
    \includegraphics[width=\linewidth]{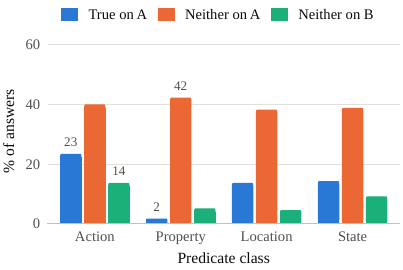}
    \caption{Answers to the target pair by predicate class (zero-shot, pooled over models).}
    \label{fig:class}
  \end{minipage}
\end{figure*}

\paragraph{Non-Maximality: Tolerance Near the Top.}
When all but one member have the property, models accept the positive definite in 25\% of cases, compared to 9\% for balanced groups and 6\% when only one member has it (repeated-measures ANOVA over models: $F(2,12) = 9.61$, $p < 0.01$).
This mirrors the non-maximal readings of plural definites in human language \citep{lasersohn-1999-pragmatic,malamud-2012-meaning,kriz-2016-homogeneity}.
Strikingly, the tolerance is one-sided: when only one member has the property, the negated definite is accepted in just 10\% of cases, barely more than for balanced groups (9\%).
The gap itself is recognized most often for maximally mixed groups (\textit{Neither} on A: 56\% vs.\ 30--34\% otherwise; $F(2,12) = 8.64$, $p < 0.01$), with no interaction with polarity ($F(2,12) = 1.19$, $p = 0.34$).

\paragraph{Predicate Semantics: The Action Penalty.}
Action predicates are accepted in 23\% of mixed scenarios (e.g., \textit{The patients improved after taking drug A}), properties in only 2\% (e.g., \textit{The hearts were red}), and locations and states in 14\% each ($F(3,18) = 7.60$, $p < 0.01$).
We hypothesize that this stems from how events are reported: descriptions such as \textit{the audience laughed} are routinely used for group events without implying that every member participated, whereas attributes such as color are checked member by member.
In contrast, predicate class has no significant affect on the use of \textit{Neither} ($F(3,18) = 0.48$, $p = 0.70$): semantics shifts which classical value models choose, but not whether they recognize the gap.

\subsection{Rationale Analysis: Truth Conditions vs.\ Truth Values}
\label{sec:rationales}

Does the Bivalence Bias reflect a failure to compute the truth conditions of plural definites, or does it arise when truth conditions are mapped onto a verdict?
We analyzed the CoT rationales, identifying whether the sentences describing the hypothesis restate it with a maximal quantifier: universal (\textit{all}, \textit{every}, \textit{each}) for the positive definite, and universal negation (\textit{none}, \textit{no}, \textit{all \dots\ not}) for the negated one.
A manual inspection of 12 random matches per group found no false positives.

\input{tables/rationales}

Table~\ref{tab:rationales} reveals that models explicitly compute homogeneous truth conditions.
Of the rationales for \textit{False} on the positive definite, 92\% restate it as a universal claim (``\textit{The hypothesis claims that all cows are in the barn. Since one cow is not in the barn, the hypothesis is false.}''), consistently across models (0.74--1.00).
For the negated definite, 48\% restate it as a universal negative (``\textit{The hypothesis states that none of the cows were in the barn.}''), systematically for \texttt{Llama-3.1} (0.94), \texttt{Mistral} (0.85), and \texttt{gemma-2} (0.76), while \texttt{Yi-1.5} (0.01) instead points to a counterexample (``\textit{Since the premise explicitly states that five of the horses are brown, the hypothesis contradicts the premise.}'').

This dissociation has a clear implication: the rationales identify exactly the conditions under which the theories of \S\ref{sec:background} predict truth (\textit{all} for the positive, \textit{none} for the negated sentence) and then treat their failure as falsity.
The Bivalence Bias is therefore not a failure of interpretation but of truth-value assignment: models map \emph{not true} onto \emph{false}.
We note that the CoT prompt mentions \textit{all}, \textit{some}, and \textit{none}, which may encourage quantified restatements, and that rationales are not guaranteed to be faithful \citep{turpin-etal-2023-language}.

\section{Conclusion}

In this work, we introduced \dataset and used it to uncover the Bivalence Bias in LLMs. Faced with a mixed group, current models almost never judge a plural definite and its negation as neither true nor false, and instead collapse the gap into falsity, especially for negated sentences.
This is not an interpretive deficit: models master the controls and compute the homogeneous truth conditions, but map their failure onto \textit{False}, and prompting does not reliably induce a third truth value.
We hope that \dataset provides a foundation for teaching LLMs the difference between \emph{not true} and \emph{false}.

\section*{Limitations and Future Work}

While our work offers insights into the Homogeneity Gap and the Bivalence Bias of LLMs, we acknowledge several limitations that define the scope of our findings and point towards future research directions.

\paragraph{Linguistic Specificity.}
Our study focuses exclusively on English.
Homogeneity has been documented beyond English, e.g., for French plural definites \citep{tieu-etal-2015-acquisition}.
It remains an open question whether the Bivalence Bias is shared across languages or reflects English-centric pre-training data.
A multilingual version of \dataset would address this question from several angles.
Models differ in the language mix of their pre-training corpora.
Evaluating them on translated items would show whether the collapse rate follows the share of a language in the training data.
Languages also differ in how they mark plural definiteness, and some lack definite articles.
Testing them would show whether the bias depends on the English form \textit{the} or on the underlying semantics.
Instruction tuning and our label set are also English-centric.
Presenting the labels in the target language and in English would separate the effect of the response format from the effect of the premise language.
Finally, a multilingual setting would test whether a rule stated in one language, as in DAP, transfers to judgments in another.

\paragraph{Ecological Validity and Syntactic Diversity.}
To strictly isolate homogeneity, we employed a template-based approach in which the premise states the composition of the group explicitly.
While this ensures high internal validity and precise minimal-pair comparisons, it sacrifices lexical and syntactic diversity: in natural discourse, group composition is often conveyed implicitly.
Future work can extend \dataset to naturalistic text, where models must infer the composition of the group from context.
This would test whether the collapse into falsity persists when models face the lexical and syntactic variation of their pre-training data.
Further extensions include collective predicates and definites embedded under quantifiers, which are central to theories of homogeneity projection \citep{kriz-chemla-2015-two}.
Varying the conversational relevance of exceptions would allow a closer study of non-maximality, which relevance modulates in humans \citep{malamud-2012-meaning,kriz-2016-homogeneity}.

\section*{Ethical Considerations}
\ifanonymous{We will release \dataset and our code under the Creative Commons Attribution 4.0 International License (CC BY 4.0).}{We release \dataset under the Creative Commons Attribution 4.0 International License (CC BY 4.0) at \url{\dataurl}.}
All models were used solely for research purposes in accordance with their licenses, and we do not redistribute any model weights.
The dataset contains no personal information or offensive content.

\paragraph{Intended Use and Risks.}
\dataset is a diagnostic dataset.
A high score does not certify a model for summarizing or verifying claims in high-stakes domains such as medicine, law, or policy.
Our premises state the group composition explicitly.
A model that passes may still fail on naturalistic text.
We see no direct risk of misuse from the data. Neither a high nor a low score should be read as a safety guarantee.

\ifanonymous{}{
\section*{Acknowledgments}
Part of this research was conducted using the Pinnacles cluster, including CENVAL-ARC compute resources (NSF \#2346744), centrally funded by the University of California, Merced, and maintained by the Cyberinfrastructure and Research Technologies (CIRT) team at UC Merced.
}

\bibliography{custom}

\appendix

\section{\texorpdfstring{\dataset}{HomogeneityNLI} Lexicon}
\label{app:lexicon}

We list the plural noun and the predicate of all 100 scenarios, grouped by predicate class.
The group size $N$ and proportion bin of each scenario are given in the released data.

\input{tables/app_lexicon}

\section{Prompts}
\label{app:prompts}

In this section, we present the prompt templates used across our four experimental conditions.
The placeholders \{premise\} and \{hypothesis\} are replaced with the corresponding fields of each item.

\begin{promptbox}{P1: Strict Logic Prompt}
\textbf{System Prompt:} You are a strict logician. Your task is to determine whether a Hypothesis is true given a Premise.
\begin{itemize}[leftmargin=*,itemsep=0pt,topsep=2pt]
  \item If the Hypothesis is true given the Premise, output "True".
  \item If the Hypothesis is false given the Premise, output "False".
  \item If the Premise makes the Hypothesis neither clearly true nor clearly false, output "Neither".
\end{itemize}
Do not use common sense assumptions. Only use the text provided.\\[3pt]
\textbf{User Input:} Premise: \{premise\} Hypothesis: \{hypothesis\}\\
Please respond with ONLY one of the following options: "True", "False", or "Neither".
\end{promptbox}

\begin{promptbox}{P2: Definition-Aware Prompt}
\textbf{System Prompt:} \dots\ [Strict Logic Instructions] \dots\\[3pt]
\textbf{Important Linguistic Rule:} Note the distinction between plural definite descriptions and quantified descriptions.
\begin{itemize}[leftmargin=*,itemsep=0pt,topsep=2pt]
  \item Plural definites (e.g., "the patients") are homogeneous: "The patients improved after taking drug A" is true only if all of them improved after taking drug A, and "The patients didn't improve after taking drug A" is true only if none of them improved after taking drug A. If some improved after taking drug A and some did not, both sentences are neither true nor false.
  \item Quantifiers (e.g., "all the patients", "not all the patients") have no such gap: if some improved after taking drug A and some did not, "All the patients improved after taking drug A" is false and "Not all the patients improved after taking drug A" is true.
\end{itemize}
Do not use common sense assumptions. Only use the text provided.\\[3pt]
\textbf{User Input:} [Same as Strict Logic Prompt]
\end{promptbox}

\begin{promptbox}{P3: Chain-of-Thought Prompt}
\textbf{System Prompt:} [Same as Strict Logic Prompt]\\[3pt]
\textbf{User Input:} Premise: \{premise\} Hypothesis: \{hypothesis\}\\
\textbf{Instruction:} First, analyze how many members of the group in the premise have the property (all, some, or none). Then analyze what the hypothesis claims about the group. Finally, provide your final label (True, False, or Neither).\\
\textbf{Output Requirement:} You must output your response in strict JSON format containing exactly two keys:
\begin{itemize}[leftmargin=*,itemsep=0pt,topsep=2pt]
  \item "reasoning": A string explaining your step-by-step logic.
  \item "label": The final judgment ("True", "False", or "Neither").
\end{itemize}
Do not include markdown formatting (like \texttt{\textasciigrave\textasciigrave\textasciigrave json}). Return ONLY the JSON object.\\
Example Format: \{"reasoning": "The premise describes a group in which...", "label": "True" / "False" / "Neither"\}
\end{promptbox}

\begin{promptbox}{P4: Enumeration Prompt}
\textbf{System Prompt:} [Same as Strict Logic Prompt]\\[3pt]
\textbf{User Input:} Premise: \{premise\} Hypothesis: \{hypothesis\}\\
\textbf{Instruction:} 1. First, go through the members of the group in the premise one by one and state for each whether the property in the hypothesis holds for it. 2. Based on this list, determine whether the hypothesis is true, false, or neither.\\
\textbf{Output Requirement:} You must output your response in strict JSON format containing exactly two keys:
\begin{itemize}[leftmargin=*,itemsep=0pt,topsep=2pt]
  \item "individuals": a list of strings, one per member of the group.
  \item "label": The final judgment ("True", "False", or "Neither").
\end{itemize}
Do not include markdown formatting (like \texttt{\textasciigrave\textasciigrave\textasciigrave json}). Return ONLY the JSON object.\\
Example Format: \{"individuals": ["..."], "label": "True" / "False" / "Neither"\}
\end{promptbox}

\section{Dataset Construction Details}
\label{app:construction}

\paragraph{Templates.}
Each scenario consists of a setup sentence introducing a group of $N$ members (e.g., ``\textit{Five hearts were drawn on the card.}''), a plural noun phrase (\textit{hearts}), and a predicate with its inflected forms.
Context sentences follow three templates: GAP: ``\textit{[k] of them [P] and [N$-$k] [did not / were not].}''; ALL: ``\textit{Every one of them [P].}''; NONE: ``\textit{None of them [P].}''
Verb agreement is adjusted for $k = 1$ and $N - k = 1$ (e.g., ``\textit{One of them was red and four were not}'').
Hypotheses follow the templates ``\textit{The [NP] [P].}'' (A, E, G), ``\textit{The [NP] [not-P].}'' with contracted negation (B, F, H), ``\textit{All the [NP] [P].}'' (C), ``\textit{Not all the [NP] [P].}'' (D), and ``\textit{Some of the [NP] [P].}'' (I, J).

\paragraph{Group Composition.}
Group sizes $N \in \{4, 5, 6, 8, 10\}$ and proportion bins are assigned cyclically within each predicate class, yielding 36 scenarios with one member having the property, 32 balanced scenarios, and 32 scenarios with all but one member having the property.
For balanced groups of odd size, $k$ is rounded up (e.g., three of five).

\paragraph{Label Variants.}
The gold label of groups A and B is \textit{Neither}.
Our generator can alternatively produce the label \textit{Unknown}; all experiments in this paper use \textit{Neither}.

\section{Implementation Details}
\label{app:implementation}

All experiments were run on single NVIDIA A100 (40GB) GPUs.
The models are open-weight, and run with the Transformers library \citep{wolf-etal-2020-transformers} (v4.57) in their native precision.
We found that Transformers v5 silently drops whitespace when encoding with the \texttt{deepseek-llm} tokenizer; all reported results therefore use v4.57.
We apply each model's chat template; for \texttt{gemma-2}, whose template does not support a system role, the system prompt is prepended to the user turn.
We disable sampling and use greedy decoding (\texttt{do\_sample=False}) to ensure deterministic outputs, and cap generations at 512 new tokens.
Generation stops at the model's end-of-sequence token or the tokenizer's end-of-turn token; the latter is added explicitly because the generation configuration of \texttt{Yi-1.5} omits it.
Labels are parsed from the \texttt{label} field of JSON outputs and otherwise from the first occurrence of \textit{True}, \textit{False}, \textit{Neither}, or \textit{Unknown} (mapped to \textit{Neither}); at most 1\% of the outputs of any run were unparseable (Table~\ref{tab:full}).
Statistical tests were computed with SciPy \citep{virtanen-etal-2020-scipy} (v1.17.1; McNemar's exact test via \texttt{binomtest} on the discordant pairs, and \texttt{chi2\_contingency}) and statsmodels \citep{seabold-perktold-2010-statsmodels} (v0.15.0; \texttt{AnovaRM} for the repeated-measures ANOVA).
Inference over all 1,000 items with the four prompts took a total of approximately 7 GPU hours across the seven models.
The models are used solely for research purposes in accordance with their original terms, and we do not redistribute any third-party model weights. Table~\ref{tab:licenses} lists the license of each model.

\begin{table}[h]
  \centering
  \small
  \begin{tabular}{@{}l l@{}}
    \toprule
    \textbf{Model} & \textbf{License} \\
    \midrule
    \texttt{Llama-3.1-8B-Instruct} & Llama 3.1 Community License \\
    \texttt{Mistral-7B-Instruct-v0.3} & Apache 2.0 \\
    \texttt{Qwen2.5-7B-Instruct} & Apache 2.0 \\
    \texttt{deepseek-llm-7b-chat} & DeepSeek License \\
    \texttt{gemma-2-9b-it} & Gemma Terms of Use \\
    \texttt{glm-4-9b-chat-hf} & GLM-4 License \\
    \texttt{Yi-1.5-9B-Chat} & Apache 2.0 \\
    \bottomrule
  \end{tabular}
  \caption{Licenses of the evaluated models.}
  \label{tab:licenses}
\end{table}

\section{Additional Results}
\label{app:results}

\paragraph{Full Results.}
Table~\ref{tab:full} reports accuracy on all ten groups.
Pooled across prompts, \textit{Neither} is predicted for at most 1.6\% of the items in any control group (C--J), confirming that models do not use it indiscriminately.

\paragraph{Response Patterns across Prompts.}
Table~\ref{tab:patterns} and Figure~\ref{fig:app-patterns} report the distribution of response patterns for all model--prompt combinations.
The Mixed pattern almost always takes the form (\textit{Neither}, \textit{False}), i.e., the gap is recognized for the positive sentence only.

\paragraph{Scalar Implicatures.}
The scalar implicature rate is 0.00 for all models in the zero-shot setting and reaches at most 0.15 (\texttt{Llama-3.1} with CoT).
Because implicatures are essentially absent, the relation between implicature computation and homogeneity predicted by \citet{magri-2014-account} cannot be estimated from our data (see Limitations).

\paragraph{Detailed Results by Proportion Bin and Predicate Class.}
Tables~\ref{tab:ratio} and~\ref{tab:class} provide the answers to the target pair for each model and prompt, broken down by the proportion of group members with the property and by predicate class.

\input{tables/app_full_results}

\input{tables/app_patterns}

\begin{figure*}[t]
  \centering
  \includegraphics[width=\textwidth]{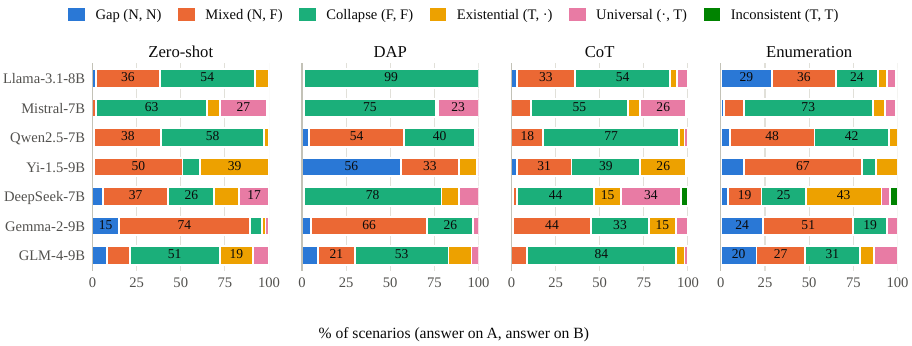}
  \caption{Response patterns on the target pair (answer on A, answer on B) for all prompting strategies, in \% of scenarios.}
  \label{fig:app-patterns}
\end{figure*}

\input{tables/app_breakdown_ratio}

\input{tables/app_breakdown_class}

\end{document}

%% file: tables/main_results.tex
\begin{table*}[t]
\centering
\small
\setlength{\tabcolsep}{4.6pt}
\begin{tabular}{llcccccccccc}
\toprule
 & & \multicolumn{5}{c}{\textbf{Accuracy ($\uparrow$)}} & \multicolumn{2}{c}{\textbf{Bias Rate}} & \textbf{HGR} & \textbf{BCR} & $\boldsymbol{\Delta_{NA}}$ \\
\cmidrule(lr){3-7} \cmidrule(lr){8-9}
\textbf{Prompt} & \textbf{Model} & \textbf{A} & \textbf{B} & \textbf{C} & \textbf{D} & \textbf{Ctrl} & \textbf{EBR} & \textbf{UBR} & ($\uparrow$) & ($\downarrow$) & ($\uparrow$) \\
\midrule
\multirow{7}{*}{\textbf{Zero-shot}} & \texttt{Llama-3.1-8B-Instruct} & \hc{22}{0.37} & \hc{2}{0.03} & \hc{60}{1.00} & \hc{60}{1.00} & \hc{60}{1.00} & \hc{5}{0.08} & \hc{0}{0.00} & \hc{1}{0.02} & \hc{32}{0.54} & \hc{60}{1.00} \\
 & \texttt{Mistral-7B-Instruct-v0.3} & \hc{1}{0.02} & \hc{0}{0.00} & \hc{60}{1.00} & \hc{60}{1.00} & \hc{60}{1.00} & \hc{5}{0.08} & \hc{17}{0.28} & \hc{0}{0.00} & \hc{38}{0.63} & \hc{43}{0.72} \\
 & \texttt{Qwen2.5-7B-Instruct} & \hc{23}{0.39} & \hc{1}{0.01} & \hc{60}{1.00} & \hc{60}{1.00} & \hc{60}{1.00} & \hc{2}{0.03} & \hc{0}{0.00} & \hc{1}{0.01} & \hc{35}{0.58} & \hc{60}{1.00} \\
 & \texttt{Yi-1.5-9B-Chat} & \hc{31}{0.51} & \hc{1}{0.02} & \hc{60}{1.00} & \hc{60}{1.00} & \hc{60}{1.00} & \hc{23}{0.39} & \hc{0}{0.00} & \hc{1}{0.01} & \hc{6}{0.10} & \hc{60}{1.00} \\
 & \texttt{deepseek-llm-7b-chat} & \hc{28}{0.46} & \hc{11}{0.18} & \hc{58}{0.96} & \hc{60}{1.00} & \hc{57}{0.95} & \hc{8}{0.14} & \hc{10}{0.17} & \hc{4}{0.06} & \hc{16}{0.26} & \hc{50}{0.83} \\
 & \texttt{gemma-2-9b-it} & \hc{50}{0.84} & \hc{13}{0.21} & \hc{60}{1.00} & \hc{59}{0.99} & \hc{60}{1.00} & \hc{1}{0.02} & \hc{1}{0.02} & \hc{9}{0.15} & \hc{4}{0.07} & \hc{58}{0.97} \\
 & \texttt{glm-4-9b-chat-hf} & \hc{12}{0.20} & \hc{7}{0.12} & \hc{60}{1.00} & \hc{60}{1.00} & \hc{60}{1.00} & \hc{11}{0.19} & \hc{5}{0.09} & \hc{5}{0.08} & \hc{31}{0.51} & \hc{55}{0.91} \\
\midrule
\multirow{7}{*}{\textbf{DAP}} & \texttt{Llama-3.1-8B-Instruct} & \hc{1}{0.01} & \hc{0}{0.00} & \hc{60}{1.00} & \hc{56}{0.94} & \hc{60}{1.00} & \hc{0}{0.00} & \hc{0}{0.00} & \hc{0}{0.00} & \hc{59}{0.99} & \hc{56}{0.94} \\
 & \texttt{Mistral-7B-Instruct-v0.3} & \hc{1}{0.01} & \hc{0}{0.00} & \hc{60}{1.00} & \hc{60}{1.00} & \hc{60}{1.00} & \hc{1}{0.01} & \hc{14}{0.23} & \hc{0}{0.00} & \hc{45}{0.75} & \hc{46}{0.77} \\
 & \texttt{Qwen2.5-7B-Instruct} & \hc{34}{0.57} & \hc{3}{0.05} & \hc{60}{1.00} & \hc{60}{1.00} & \hc{60}{1.00} & \hc{1}{0.01} & \hc{1}{0.01} & \hc{2}{0.04} & \hc{24}{0.40} & \hc{59}{0.99} \\
 & \texttt{Yi-1.5-9B-Chat} & \hc{54}{0.90} & \hc{37}{0.61} & \hc{60}{1.00} & \hc{60}{1.00} & \hc{60}{0.99} & \hc{6}{0.10} & \hc{1}{0.01} & \hc{34}{0.56} & \hc{0}{0.00} & \hc{59}{0.99} \\
 & \texttt{deepseek-llm-7b-chat} & \hc{1}{0.02} & \hc{0}{0.00} & \hc{60}{1.00} & \hc{60}{1.00} & \hc{60}{0.99} & \hc{6}{0.10} & \hc{7}{0.11} & \hc{0}{0.00} & \hc{47}{0.78} & \hc{53}{0.89} \\
 & \texttt{gemma-2-9b-it} & \hc{37}{0.61} & \hc{9}{0.15} & \hc{60}{1.00} & \hc{60}{1.00} & \hc{60}{1.00} & \hc{0}{0.00} & \hc{2}{0.03} & \hc{3}{0.05} & \hc{16}{0.26} & \hc{58}{0.97} \\
 & \texttt{glm-4-9b-chat-hf} & \hc{18}{0.30} & \hc{7}{0.12} & \hc{60}{1.00} & \hc{60}{1.00} & \hc{60}{1.00} & \hc{8}{0.13} & \hc{2}{0.04} & \hc{5}{0.09} & \hc{32}{0.53} & \hc{58}{0.96} \\
\midrule
\multirow{7}{*}{\textbf{CoT}} & \texttt{Llama-3.1-8B-Instruct} & \hc{12}{0.20} & \hc{14}{0.23} & \hc{44}{0.74} & \hc{55}{0.92} & \hc{60}{1.00} & \hc{2}{0.04} & \hc{4}{0.06} & \hc{2}{0.03} & \hc{32}{0.54} & \hc{52}{0.86} \\
 & \texttt{Mistral-7B-Instruct-v0.3} & \hc{1}{0.01} & \hc{7}{0.11} & \hc{60}{1.00} & \hc{55}{0.92} & \hc{60}{0.99} & \hc{5}{0.08} & \hc{16}{0.27} & \hc{0}{0.00} & \hc{33}{0.55} & \hc{39}{0.65} \\
 & \texttt{Qwen2.5-7B-Instruct} & \hc{10}{0.17} & \hc{1}{0.01} & \hc{60}{1.00} & \hc{60}{1.00} & \hc{60}{1.00} & \hc{2}{0.03} & \hc{1}{0.02} & \hc{0}{0.00} & \hc{46}{0.77} & \hc{59}{0.98} \\
 & \texttt{Yi-1.5-9B-Chat} & \hc{17}{0.29} & \hc{5}{0.09} & \hc{60}{1.00} & \hc{60}{1.00} & \hc{60}{1.00} & \hc{16}{0.26} & \hc{1}{0.01} & \hc{2}{0.03} & \hc{23}{0.39} & \hc{59}{0.99} \\
 & \texttt{deepseek-llm-7b-chat} & \hc{2}{0.04} & \hc{2}{0.04} & \hc{58}{0.96} & \hc{44}{0.74} & \hc{60}{0.99} & \hc{11}{0.19} & \hc{23}{0.38} & \hc{1}{0.01} & \hc{26}{0.44} & \hc{22}{0.36} \\
 & \texttt{gemma-2-9b-it} & \hc{26}{0.43} & \hc{2}{0.03} & \hc{60}{1.00} & \hc{60}{1.00} & \hc{60}{1.00} & \hc{9}{0.15} & \hc{4}{0.07} & \hc{1}{0.01} & \hc{20}{0.33} & \hc{56}{0.93} \\
 & \texttt{glm-4-9b-chat-hf} & \hc{5}{0.09} & \hc{0}{0.00} & \hc{59}{0.98} & \hc{58}{0.96} & \hc{60}{1.00} & \hc{3}{0.05} & \hc{1}{0.02} & \hc{0}{0.00} & \hc{50}{0.84} & \hc{56}{0.94} \\
\midrule
\multirow{7}{*}{\textbf{Enumeration}} & \texttt{Llama-3.1-8B-Instruct} & \hc{40}{0.67} & \hc{18}{0.30} & \hc{58}{0.96} & \hc{58}{0.97} & \hc{48}{0.79} & \hc{4}{0.06} & \hc{4}{0.06} & \hc{17}{0.29} & \hc{14}{0.24} & \hc{55}{0.91} \\
 & \texttt{Mistral-7B-Instruct-v0.3} & \hc{6}{0.10} & \hc{4}{0.06} & \hc{60}{1.00} & \hc{49}{0.81} & \hc{60}{1.00} & \hc{5}{0.08} & \hc{4}{0.07} & \hc{1}{0.02} & \hc{44}{0.73} & \hc{44}{0.74} \\
 & \texttt{Qwen2.5-7B-Instruct} & \hc{31}{0.51} & \hc{4}{0.07} & \hc{60}{1.00} & \hc{59}{0.99} & \hc{60}{1.00} & \hc{3}{0.05} & \hc{0}{0.00} & \hc{3}{0.05} & \hc{25}{0.42} & \hc{59}{0.99} \\
 & \texttt{Yi-1.5-9B-Chat} & \hc{46}{0.77} & \hc{10}{0.17} & \hc{58}{0.96} & \hc{60}{1.00} & \hc{60}{1.00} & \hc{7}{0.12} & \hc{0}{0.00} & \hc{8}{0.13} & \hc{5}{0.08} & \hc{60}{1.00} \\
 & \texttt{deepseek-llm-7b-chat}$^\dagger$ & \hc{10}{0.16} & \hc{15}{0.25} & \hc{34}{0.56} & \hc{7}{0.11} & \hc{37}{0.62} & \hc{28}{0.47} & \hc{5}{0.09} & \hc{2}{0.04} & \hc{15}{0.25} & \hc{1}{0.02} \\
 & \texttt{gemma-2-9b-it} & \hc{47}{0.78} & \hc{16}{0.26} & \hc{60}{1.00} & \hc{60}{1.00} & \hc{60}{1.00} & \hc{0}{0.00} & \hc{4}{0.06} & \hc{14}{0.24} & \hc{11}{0.19} & \hc{56}{0.94} \\
 & \texttt{glm-4-9b-chat-hf} & \hc{27}{0.45} & \hc{16}{0.27} & \hc{59}{0.99} & \hc{59}{0.99} & \hc{60}{1.00} & \hc{5}{0.08} & \hc{8}{0.13} & \hc{12}{0.20} & \hc{19}{0.31} & \hc{52}{0.86} \\
\bottomrule
\end{tabular}
\caption{Model performance under the four prompting strategies: accuracy on groups A--D and the uniform-context controls (Ctrl: E--H), Existential and Universal Bias Rates, Homogeneity Gap Rate (HGR), Bivalent Collapse Rate (BCR), and Negation Asymmetry Gap ($\Delta_{NA}$). $^\dagger$Degenerate run (\S\ref{sec:prompting}).}
\label{tab:main}
\end{table*}

%% file: tables/rationales.tex
\begin{table}[t]
\centering
\small
\setlength{\tabcolsep}{5pt}
\begin{tabular}{lcc}
\toprule
\textbf{Model} & \textbf{A: ``all''} & \textbf{B: ``none''} \\
\midrule
\texttt{Llama-3.1-8B} & 1.00 {\scriptsize (76/76)} & 0.94 {\scriptsize (67/71)} \\
\texttt{Mistral-7B} & 0.98 {\scriptsize (89/91)} & 0.85 {\scriptsize (53/62)} \\
\texttt{Qwen2.5-7B} & 0.74 {\scriptsize (59/80)} & 0.34 {\scriptsize (33/97)} \\
\texttt{Yi-1.5-9B} & 0.93 {\scriptsize (42/45)} & 0.01 {\scriptsize (1/90)} \\
\texttt{DeepSeek-7B} & 0.91 {\scriptsize (70/77)} & 0.22 {\scriptsize (13/58)} \\
\texttt{Gemma-2-9B} & 0.98 {\scriptsize (41/42)} & 0.76 {\scriptsize (68/90)} \\
\texttt{GLM-4-9B} & 0.93 {\scriptsize (80/86)} & 0.37 {\scriptsize (36/98)} \\
\midrule
\textbf{All} & 0.92 {\scriptsize (457/497)} & 0.48 {\scriptsize (271/566)} \\
\bottomrule
\end{tabular}
\caption{Share of CoT rationales for \textit{False} that restate the hypothesis as \textit{all} (A) or \textit{none} (B).}
\label{tab:rationales}
\end{table}

%% file: tables/app_lexicon.tex
\paragraph{Action (25).} \textit{kids}: laughed; \textit{students}: passed the exam; \textit{runners}: finished the race; \textit{guests}: showed up; \textit{dogs}: barked; \textit{children}: tasted the broccoli; \textit{volunteers}: attended the training; \textit{employees}: responded; \textit{players}: scored a goal; \textit{birds}: flew away; \textit{passengers}: complained; \textit{actors}: remembered their lines; \textit{candidates}: were interviewed; \textit{babies}: cried; \textit{swimmers}: reached the island; \textit{neighbors}: replied; \textit{cats}: ate it; \textit{climbers}: reached the summit; \textit{tourists}: bought souvenirs; \textit{students}: submitted it on time; \textit{kids}: jumped into the pool; \textit{patients}: developed a fever; \textit{musicians}: agreed; \textit{hikers}: turned back early; \textit{jurors}: took notes.

\paragraph{Property (25).} \textit{trucks}: were blue; \textit{hearts}: were red; \textit{balloons}: were yellow; \textit{umbrellas}: were striped; \textit{cups}: were made of glass; \textit{letters}: were handwritten; \textit{tents}: were orange; \textit{hats}: were made of wool; \textit{coins}: were made of gold; \textit{cars}: were electric; \textit{bouquets}: were white; \textit{shirts}: were green; \textit{doors}: were painted black; \textit{books}: were signed by the author; \textit{chairs}: were wooden; \textit{cakes}: were chocolate; \textit{sheep}: were black; \textit{horses}: were brown; \textit{kittens}: were female; \textit{tests}: were written in pencil; \textit{stars}: were silver; \textit{vases}: were made of porcelain; \textit{pillows}: were square; \textit{tulips}: were purple; \textit{bikes}: were fitted with baskets.

\paragraph{Location (25).} \textit{toy cars}: were in the garage; \textit{knives}: were in the drawer; \textit{cows}: were in the barn; \textit{suitcases}: were in the car; \textit{buses}: were parked at the depot; \textit{printers}: were on the second floor; \textit{bicycles}: were in the shed; \textit{children}: were hiding in the attic; \textit{trophies}: were in the display case; \textit{crates of wine}: were in the cellar; \textit{ducks}: were in the pond; \textit{fingerprints}: were on the window; \textit{books}: were in his backpack; \textit{ambulances}: were at the main entrance; \textit{paintings}: were in storage; \textit{boats}: were in the harbor; \textit{plates}: were in the dishwasher; \textit{penguins}: were in the water; \textit{laptops}: were on the charging cart; \textit{planes}: were in the hangar; \textit{scarves}: were in the closet; \textit{brushes}: were in the jar; \textit{guests}: were in their rooms; \textit{plants}: were on the balcony; \textit{keys}: were on the hook.

\paragraph{State (25).} \textit{puppies}: were asleep; \textit{windows}: were open; \textit{lamps}: were switched on; \textit{doors}: were locked; \textit{towels}: were dry; \textit{candles}: were lit; \textit{glasses}: were full; \textit{phones}: were charged; \textit{ovens}: were hot; \textit{bananas}: were ripe; \textit{hamsters}: were awake; \textit{parking spots}: were empty; \textit{rooms}: were occupied; \textit{pipes}: were frozen; \textit{snowmen}: were melting; \textit{players}: were injured; \textit{computers}: were running; \textit{faucets}: were dripping; \textit{cartons of milk}: were expired; \textit{fields}: were flooded; \textit{gates}: were closed; \textit{traffic lights}: were broken; \textit{patients}: were feverish; \textit{bottles}: were cold; \textit{soldiers}: were on duty.

%% file: tables/app_full_results.tex
\begin{table*}[t]
\centering
\small
\setlength{\tabcolsep}{3.4pt}
\begin{tabular}{llcccccccccccc}
\toprule
\textbf{Prompt} & \textbf{Model} & \textbf{A} & \textbf{B} & \textbf{C} & \textbf{D} & \textbf{E} & \textbf{F} & \textbf{G} & \textbf{H} & \textbf{I} & \textbf{J} & \textbf{SI} & \textbf{Inv.} \\
\midrule
\multirow{7}{*}{\textbf{Zero-shot}} & \texttt{Llama-3.1-8B-Instruct} & 0.37 & 0.03 & 1.00 & 1.00 & 1.00 & 1.00 & 1.00 & 1.00 & 1.00 & 1.00 & 0.00 & 0.00 \\
 & \texttt{Mistral-7B-Instruct-v0.3} & 0.02 & 0.00 & 1.00 & 1.00 & 1.00 & 1.00 & 1.00 & 1.00 & 1.00 & 1.00 & 0.00 & 0.00 \\
 & \texttt{Qwen2.5-7B-Instruct} & 0.39 & 0.01 & 1.00 & 1.00 & 1.00 & 1.00 & 1.00 & 1.00 & 1.00 & 1.00 & 0.00 & 0.00 \\
 & \texttt{Yi-1.5-9B-Chat} & 0.51 & 0.02 & 1.00 & 1.00 & 1.00 & 1.00 & 1.00 & 1.00 & 1.00 & 1.00 & 0.00 & 0.00 \\
 & \texttt{deepseek-llm-7b-chat} & 0.46 & 0.18 & 0.96 & 1.00 & 1.00 & 1.00 & 0.82 & 1.00 & 1.00 & 1.00 & 0.00 & 0.00 \\
 & \texttt{gemma-2-9b-it} & 0.84 & 0.21 & 1.00 & 0.99 & 1.00 & 1.00 & 1.00 & 1.00 & 1.00 & 1.00 & 0.00 & 0.00 \\
 & \texttt{glm-4-9b-chat-hf} & 0.20 & 0.12 & 1.00 & 1.00 & 1.00 & 1.00 & 1.00 & 1.00 & 1.00 & 1.00 & 0.00 & 0.00 \\
\midrule
\multirow{7}{*}{\textbf{DAP}} & \texttt{Llama-3.1-8B-Instruct} & 0.01 & 0.00 & 1.00 & 0.94 & 1.00 & 1.00 & 1.00 & 1.00 & 1.00 & 1.00 & 0.00 & 0.00 \\
 & \texttt{Mistral-7B-Instruct-v0.3} & 0.01 & 0.00 & 1.00 & 1.00 & 1.00 & 1.00 & 1.00 & 1.00 & 1.00 & 1.00 & 0.00 & 0.00 \\
 & \texttt{Qwen2.5-7B-Instruct} & 0.57 & 0.05 & 1.00 & 1.00 & 1.00 & 1.00 & 1.00 & 1.00 & 1.00 & 1.00 & 0.00 & 0.00 \\
 & \texttt{Yi-1.5-9B-Chat} & 0.90 & 0.61 & 1.00 & 1.00 & 1.00 & 0.99 & 0.99 & 1.00 & 1.00 & 1.00 & 0.00 & 0.00 \\
 & \texttt{deepseek-llm-7b-chat} & 0.02 & 0.00 & 1.00 & 1.00 & 1.00 & 1.00 & 0.97 & 1.00 & 1.00 & 1.00 & 0.00 & 0.00 \\
 & \texttt{gemma-2-9b-it} & 0.61 & 0.15 & 1.00 & 1.00 & 1.00 & 1.00 & 1.00 & 1.00 & 0.99 & 1.00 & 0.01 & 0.00 \\
 & \texttt{glm-4-9b-chat-hf} & 0.30 & 0.12 & 1.00 & 1.00 & 1.00 & 1.00 & 1.00 & 1.00 & 1.00 & 1.00 & 0.00 & 0.00 \\
\midrule
\multirow{7}{*}{\textbf{CoT}} & \texttt{Llama-3.1-8B-Instruct} & 0.20 & 0.23 & 0.74 & 0.92 & 1.00 & 1.00 & 0.99 & 1.00 & 0.85 & 0.99 & 0.15 & 0.00 \\
 & \texttt{Mistral-7B-Instruct-v0.3} & 0.01 & 0.11 & 1.00 & 0.92 & 1.00 & 1.00 & 0.98 & 1.00 & 0.96 & 1.00 & 0.04 & 0.00 \\
 & \texttt{Qwen2.5-7B-Instruct} & 0.17 & 0.01 & 1.00 & 1.00 & 1.00 & 1.00 & 1.00 & 1.00 & 1.00 & 1.00 & 0.00 & 0.00 \\
 & \texttt{Yi-1.5-9B-Chat} & 0.29 & 0.09 & 1.00 & 1.00 & 1.00 & 1.00 & 1.00 & 1.00 & 1.00 & 1.00 & 0.00 & 0.00 \\
 & \texttt{deepseek-llm-7b-chat} & 0.04 & 0.04 & 0.96 & 0.74 & 1.00 & 0.98 & 1.00 & 1.00 & 0.86 & 1.00 & 0.14 & 0.00 \\
 & \texttt{gemma-2-9b-it} & 0.43 & 0.03 & 1.00 & 1.00 & 1.00 & 1.00 & 1.00 & 1.00 & 1.00 & 1.00 & 0.00 & 0.00 \\
 & \texttt{glm-4-9b-chat-hf} & 0.09 & 0.00 & 0.98 & 0.96 & 1.00 & 1.00 & 1.00 & 1.00 & 0.96 & 1.00 & 0.04 & 0.00 \\
\midrule
\multirow{7}{*}{\textbf{Enumeration}} & \texttt{Llama-3.1-8B-Instruct} & 0.67 & 0.30 & 0.96 & 0.97 & 0.67 & 0.88 & 0.83 & 0.79 & 0.98 & 0.99 & 0.02 & 0.00 \\
 & \texttt{Mistral-7B-Instruct-v0.3} & 0.10 & 0.06 & 1.00 & 0.81 & 1.00 & 1.00 & 1.00 & 1.00 & 1.00 & 0.99 & 0.00 & 0.00 \\
 & \texttt{Qwen2.5-7B-Instruct} & 0.51 & 0.07 & 1.00 & 0.99 & 1.00 & 1.00 & 1.00 & 0.99 & 1.00 & 0.99 & 0.00 & 0.00 \\
 & \texttt{Yi-1.5-9B-Chat} & 0.77 & 0.17 & 0.96 & 1.00 & 1.00 & 1.00 & 1.00 & 1.00 & 1.00 & 1.00 & 0.00 & 0.00 \\
 & \texttt{deepseek-llm-7b-chat}$^\dagger$ & 0.16 & 0.25 & 0.56 & 0.11 & 0.96 & 0.89 & 0.39 & 0.25 & 0.95 & 1.00 & 0.05 & 0.00 \\
 & \texttt{gemma-2-9b-it} & 0.78 & 0.26 & 1.00 & 1.00 & 1.00 & 1.00 & 1.00 & 1.00 & 1.00 & 1.00 & 0.00 & 0.00 \\
 & \texttt{glm-4-9b-chat-hf} & 0.45 & 0.27 & 0.99 & 0.99 & 1.00 & 1.00 & 0.99 & 1.00 & 1.00 & 0.99 & 0.00 & 0.01 \\
\bottomrule
\end{tabular}
\caption{Accuracy on all ten groups of \textsc{HomogeneityNLI}, with the scalar implicature rate (SI: share of group-I items not judged \textit{True}) and the share of unparseable outputs (Inv.). Gold labels: A, B = \textit{Neither}; C, F, G = \textit{False}; D, E, H, I, J = \textit{True} (group I is \textit{True} under the literal reading of \textit{some}).}
\label{tab:full}
\end{table*}

%% file: tables/app_patterns.tex
\begin{table*}[t]
\centering
\small
\setlength{\tabcolsep}{4pt}
\begin{tabular}{llccccccccc}
\toprule
 & & \textbf{Gap} & \multicolumn{2}{c}{\textbf{Mixed}} & \textbf{Collapse} & \textbf{Exist.} & \textbf{Univ.} & \textbf{Incons.} & \textbf{Reject} & \textbf{Ctrl} \\
\textbf{Prompt} & \textbf{Model} & (N,N) & (N,F) & (F,N) & (F,F) & (T,$\cdot$) & ($\cdot$,T) & (T,T) & both & OK \\
\midrule
\multirow{7}{*}{\textbf{Zero-shot}} & \texttt{Llama-3.1-8B-Instruct} & 0.02 & 0.35 & 0.01 & 0.54 & 0.08 & 0.00 & 0.00 & 0.92 & 100 \\
 & \texttt{Mistral-7B-Instruct-v0.3} & 0.00 & 0.02 & 0.00 & 0.63 & 0.07 & 0.27 & 0.01 & 0.65 & 100 \\
 & \texttt{Qwen2.5-7B-Instruct} & 0.01 & 0.38 & 0.00 & 0.58 & 0.03 & 0.00 & 0.00 & 0.97 & 100 \\
 & \texttt{Yi-1.5-9B-Chat} & 0.01 & 0.50 & 0.00 & 0.10 & 0.39 & 0.00 & 0.00 & 0.61 & 100 \\
 & \texttt{deepseek-llm-7b-chat} & 0.06 & 0.30 & 0.07 & 0.26 & 0.14 & 0.17 & 0.00 & 0.69 & 82 \\
 & \texttt{gemma-2-9b-it} & 0.15 & 0.68 & 0.06 & 0.07 & 0.02 & 0.02 & 0.00 & 0.96 & 100 \\
 & \texttt{glm-4-9b-chat-hf} & 0.08 & 0.09 & 0.04 & 0.51 & 0.19 & 0.09 & 0.00 & 0.72 & 100 \\
\midrule
\multirow{7}{*}{\textbf{DAP}} & \texttt{Llama-3.1-8B-Instruct} & 0.00 & 0.01 & 0.00 & 0.99 & 0.00 & 0.00 & 0.00 & 1.00 & 100 \\
 & \texttt{Mistral-7B-Instruct-v0.3} & 0.00 & 0.01 & 0.00 & 0.75 & 0.01 & 0.23 & 0.00 & 0.76 & 100 \\
 & \texttt{Qwen2.5-7B-Instruct} & 0.04 & 0.53 & 0.01 & 0.40 & 0.01 & 0.01 & 0.00 & 0.98 & 100 \\
 & \texttt{Yi-1.5-9B-Chat} & 0.56 & 0.33 & 0.00 & 0.00 & 0.10 & 0.01 & 0.00 & 0.89 & 98 \\
 & \texttt{deepseek-llm-7b-chat} & 0.00 & 0.01 & 0.00 & 0.78 & 0.10 & 0.11 & 0.00 & 0.79 & 97 \\
 & \texttt{gemma-2-9b-it} & 0.05 & 0.56 & 0.10 & 0.26 & 0.00 & 0.03 & 0.00 & 0.97 & 100 \\
 & \texttt{glm-4-9b-chat-hf} & 0.09 & 0.21 & 0.00 & 0.53 & 0.13 & 0.04 & 0.00 & 0.83 & 100 \\
\midrule
\multirow{7}{*}{\textbf{CoT}} & \texttt{Llama-3.1-8B-Instruct} & 0.03 & 0.14 & 0.19 & 0.54 & 0.04 & 0.06 & 0.00 & 0.90 & 99 \\
 & \texttt{Mistral-7B-Instruct-v0.3} & 0.00 & 0.01 & 0.10 & 0.55 & 0.07 & 0.26 & 0.01 & 0.66 & 98 \\
 & \texttt{Qwen2.5-7B-Instruct} & 0.00 & 0.17 & 0.01 & 0.77 & 0.03 & 0.02 & 0.00 & 0.95 & 100 \\
 & \texttt{Yi-1.5-9B-Chat} & 0.03 & 0.26 & 0.05 & 0.39 & 0.26 & 0.01 & 0.00 & 0.73 & 100 \\
 & \texttt{deepseek-llm-7b-chat} & 0.01 & 0.00 & 0.02 & 0.44 & 0.15 & 0.34 & 0.04 & 0.47 & 98 \\
 & \texttt{gemma-2-9b-it} & 0.01 & 0.42 & 0.02 & 0.33 & 0.15 & 0.07 & 0.00 & 0.78 & 100 \\
 & \texttt{glm-4-9b-chat-hf} & 0.00 & 0.09 & 0.00 & 0.84 & 0.05 & 0.02 & 0.00 & 0.93 & 100 \\
\midrule
\multirow{7}{*}{\textbf{Enumeration}} & \texttt{Llama-3.1-8B-Instruct} & 0.29 & 0.35 & 0.01 & 0.24 & 0.05 & 0.05 & 0.01 & 0.89 & 58 \\
 & \texttt{Mistral-7B-Instruct-v0.3} & 0.02 & 0.07 & 0.04 & 0.73 & 0.07 & 0.06 & 0.01 & 0.86 & 100 \\
 & \texttt{Qwen2.5-7B-Instruct} & 0.05 & 0.46 & 0.02 & 0.42 & 0.05 & 0.00 & 0.00 & 0.95 & 99 \\
 & \texttt{Yi-1.5-9B-Chat} & 0.13 & 0.64 & 0.03 & 0.08 & 0.12 & 0.00 & 0.00 & 0.88 & 100 \\
 & \texttt{deepseek-llm-7b-chat}$^\dagger$ & 0.04 & 0.11 & 0.08 & 0.25 & 0.43 & 0.05 & 0.04 & 0.48 & 10 \\
 & \texttt{gemma-2-9b-it} & 0.24 & 0.49 & 0.02 & 0.19 & 0.00 & 0.06 & 0.00 & 0.94 & 100 \\
 & \texttt{glm-4-9b-chat-hf} & 0.20 & 0.21 & 0.06 & 0.31 & 0.08 & 0.13 & 0.00 & 0.79 & 98 \\
\bottomrule
\end{tabular}
\caption{Distribution of per-scenario response patterns on the target pair (answer on A, answer on B), following the participant classification of \citet{tieu-etal-2015-acquisition}. N = \textit{Neither}, F = \textit{False}, T = \textit{True}. \textit{Reject both} (= Gap + Mixed + Collapse) is the ``homogeneous'' category of a binary truth-value judgment task. Ctrl-OK: number of scenarios (of 100) whose four definite controls (E--H) are all correct.}
\label{tab:patterns}
\end{table*}

%% file: tables/app_breakdown_ratio.tex
\begin{table*}[t]
\centering
\small
\setlength{\tabcolsep}{3.6pt}
\begin{tabular}{llccccccccc}
\toprule
 & & \multicolumn{3}{c}{\textbf{1 of N}} & \multicolumn{3}{c}{\textbf{About half}} & \multicolumn{3}{c}{\textbf{N$-$1 of N}} \\
\cmidrule(lr){3-5} \cmidrule(lr){6-8} \cmidrule(lr){9-11}
\textbf{Prompt} & \textbf{Model} & T$_A$ & N$_A$ & N$_B$ & T$_A$ & N$_A$ & N$_B$ & T$_A$ & N$_A$ & N$_B$ \\
\midrule
\multirow{7}{*}{\textbf{Zero-shot}} & \texttt{Llama-3.1-8B-Instruct} & 0.00 & 0.33 & 0.03 & 0.06 & 0.69 & 0.06 & 0.19 & 0.09 & 0.00 \\
 & \texttt{Mistral-7B-Instruct-v0.3} & 0.06 & 0.00 & 0.00 & 0.03 & 0.03 & 0.00 & 0.16 & 0.03 & 0.00 \\
 & \texttt{Qwen2.5-7B-Instruct} & 0.00 & 0.06 & 0.00 & 0.03 & 0.72 & 0.03 & 0.06 & 0.44 & 0.00 \\
 & \texttt{Yi-1.5-9B-Chat} & 0.28 & 0.53 & 0.00 & 0.38 & 0.62 & 0.06 & 0.53 & 0.38 & 0.00 \\
 & \texttt{deepseek-llm-7b-chat} & 0.06 & 0.64 & 0.08 & 0.06 & 0.53 & 0.22 & 0.31 & 0.19 & 0.25 \\
 & \texttt{gemma-2-9b-it} & 0.03 & 0.64 & 0.39 & 0.00 & 0.97 & 0.22 & 0.03 & 0.94 & 0.00 \\
 & \texttt{glm-4-9b-chat-hf} & 0.00 & 0.19 & 0.06 & 0.09 & 0.38 & 0.31 & 0.50 & 0.03 & 0.00 \\
\midrule
\multirow{7}{*}{\textbf{DAP}} & \texttt{Llama-3.1-8B-Instruct} & 0.00 & 0.00 & 0.00 & 0.00 & 0.03 & 0.00 & 0.00 & 0.00 & 0.00 \\
 & \texttt{Mistral-7B-Instruct-v0.3} & 0.00 & 0.00 & 0.00 & 0.00 & 0.03 & 0.00 & 0.03 & 0.00 & 0.00 \\
 & \texttt{Qwen2.5-7B-Instruct} & 0.00 & 0.14 & 0.03 & 0.00 & 0.94 & 0.12 & 0.03 & 0.69 & 0.00 \\
 & \texttt{Yi-1.5-9B-Chat} & 0.08 & 0.92 & 0.44 & 0.09 & 0.91 & 0.84 & 0.12 & 0.88 & 0.56 \\
 & \texttt{deepseek-llm-7b-chat} & 0.06 & 0.06 & 0.00 & 0.06 & 0.00 & 0.00 & 0.19 & 0.00 & 0.00 \\
 & \texttt{gemma-2-9b-it} & 0.00 & 0.31 & 0.31 & 0.00 & 0.81 & 0.12 & 0.00 & 0.75 & 0.00 \\
 & \texttt{glm-4-9b-chat-hf} & 0.00 & 0.17 & 0.08 & 0.09 & 0.56 & 0.28 & 0.31 & 0.19 & 0.00 \\
\midrule
\multirow{7}{*}{\textbf{CoT}} & \texttt{Llama-3.1-8B-Instruct} & 0.03 & 0.17 & 0.14 & 0.03 & 0.31 & 0.31 & 0.06 & 0.12 & 0.25 \\
 & \texttt{Mistral-7B-Instruct-v0.3} & 0.11 & 0.00 & 0.06 & 0.06 & 0.03 & 0.19 & 0.06 & 0.00 & 0.09 \\
 & \texttt{Qwen2.5-7B-Instruct} & 0.03 & 0.08 & 0.00 & 0.00 & 0.22 & 0.03 & 0.06 & 0.22 & 0.00 \\
 & \texttt{Yi-1.5-9B-Chat} & 0.14 & 0.44 & 0.06 & 0.31 & 0.22 & 0.16 & 0.34 & 0.19 & 0.06 \\
 & \texttt{deepseek-llm-7b-chat} & 0.19 & 0.00 & 0.03 & 0.22 & 0.09 & 0.06 & 0.16 & 0.03 & 0.03 \\
 & \texttt{gemma-2-9b-it} & 0.11 & 0.42 & 0.03 & 0.12 & 0.47 & 0.03 & 0.22 & 0.41 & 0.03 \\
 & \texttt{glm-4-9b-chat-hf} & 0.00 & 0.03 & 0.00 & 0.09 & 0.09 & 0.00 & 0.06 & 0.16 & 0.00 \\
\midrule
\multirow{7}{*}{\textbf{Enumeration}} & \texttt{Llama-3.1-8B-Instruct} & 0.00 & 0.69 & 0.36 & 0.06 & 0.81 & 0.22 & 0.12 & 0.50 & 0.31 \\
 & \texttt{Mistral-7B-Instruct-v0.3} & 0.00 & 0.00 & 0.08 & 0.12 & 0.19 & 0.09 & 0.12 & 0.12 & 0.00 \\
 & \texttt{Qwen2.5-7B-Instruct} & 0.00 & 0.22 & 0.06 & 0.06 & 0.88 & 0.16 & 0.09 & 0.47 & 0.00 \\
 & \texttt{Yi-1.5-9B-Chat} & 0.06 & 0.72 & 0.11 & 0.09 & 0.91 & 0.34 & 0.22 & 0.69 & 0.06 \\
 & \texttt{deepseek-llm-7b-chat}$^\dagger$ & 0.28 & 0.31 & 0.25 & 0.69 & 0.03 & 0.16 & 0.47 & 0.12 & 0.34 \\
 & \texttt{gemma-2-9b-it} & 0.00 & 0.64 & 0.19 & 0.00 & 1.00 & 0.44 & 0.00 & 0.72 & 0.16 \\
 & \texttt{glm-4-9b-chat-hf} & 0.00 & 0.36 & 0.17 & 0.03 & 0.81 & 0.53 & 0.22 & 0.19 & 0.12 \\
\bottomrule
\end{tabular}
\caption{Breakdown by the proportion of group members with the property (GAP contexts): share of \textit{True} on A (T$_A$), \textit{Neither} on A (N$_A$), and \textit{Neither} on B (N$_B$).}
\label{tab:ratio}
\end{table*}

%% file: tables/app_breakdown_class.tex
\begin{table*}[t]
\centering
\small
\setlength{\tabcolsep}{3.6pt}
\begin{tabular}{llcccccccccccc}
\toprule
 & & \multicolumn{3}{c}{\textbf{Action}} & \multicolumn{3}{c}{\textbf{Property}} & \multicolumn{3}{c}{\textbf{Location}} & \multicolumn{3}{c}{\textbf{State}} \\
\cmidrule(lr){3-5} \cmidrule(lr){6-8} \cmidrule(lr){9-11} \cmidrule(lr){12-14}
\textbf{Prompt} & \textbf{Model} & T$_A$ & N$_A$ & N$_B$ & T$_A$ & N$_A$ & N$_B$ & T$_A$ & N$_A$ & N$_B$ & T$_A$ & N$_A$ & N$_B$ \\
\midrule
\multirow{7}{*}{\textbf{Zero-shot}} & \texttt{Llama-3.1-8B-Instruct} & 0.08 & 0.24 & 0.04 & 0.00 & 0.32 & 0.08 & 0.08 & 0.44 & 0.00 & 0.16 & 0.48 & 0.00 \\
 & \texttt{Mistral-7B-Instruct-v0.3} & 0.24 & 0.00 & 0.00 & 0.00 & 0.00 & 0.00 & 0.04 & 0.08 & 0.00 & 0.04 & 0.00 & 0.00 \\
 & \texttt{Qwen2.5-7B-Instruct} & 0.04 & 0.60 & 0.00 & 0.00 & 0.16 & 0.00 & 0.04 & 0.40 & 0.04 & 0.04 & 0.40 & 0.00 \\
 & \texttt{Yi-1.5-9B-Chat} & 0.60 & 0.32 & 0.08 & 0.12 & 0.84 & 0.00 & 0.52 & 0.32 & 0.00 & 0.32 & 0.56 & 0.00 \\
 & \texttt{deepseek-llm-7b-chat} & 0.28 & 0.36 & 0.48 & 0.00 & 0.64 & 0.00 & 0.12 & 0.36 & 0.00 & 0.16 & 0.48 & 0.24 \\
 & \texttt{gemma-2-9b-it} & 0.04 & 0.88 & 0.12 & 0.00 & 0.84 & 0.20 & 0.00 & 0.96 & 0.16 & 0.04 & 0.68 & 0.36 \\
 & \texttt{glm-4-9b-chat-hf} & 0.36 & 0.40 & 0.24 & 0.00 & 0.16 & 0.08 & 0.16 & 0.12 & 0.12 & 0.24 & 0.12 & 0.04 \\
\midrule
\multirow{7}{*}{\textbf{DAP}} & \texttt{Llama-3.1-8B-Instruct} & 0.00 & 0.00 & 0.00 & 0.00 & 0.00 & 0.00 & 0.00 & 0.04 & 0.00 & 0.00 & 0.00 & 0.00 \\
 & \texttt{Mistral-7B-Instruct-v0.3} & 0.04 & 0.00 & 0.00 & 0.00 & 0.00 & 0.00 & 0.00 & 0.04 & 0.00 & 0.00 & 0.00 & 0.00 \\
 & \texttt{Qwen2.5-7B-Instruct} & 0.00 & 0.76 & 0.04 & 0.00 & 0.36 & 0.00 & 0.00 & 0.68 & 0.12 & 0.04 & 0.48 & 0.04 \\
 & \texttt{Yi-1.5-9B-Chat} & 0.12 & 0.88 & 0.88 & 0.04 & 0.96 & 0.72 & 0.16 & 0.84 & 0.20 & 0.08 & 0.92 & 0.64 \\
 & \texttt{deepseek-llm-7b-chat} & 0.12 & 0.00 & 0.00 & 0.00 & 0.00 & 0.00 & 0.16 & 0.00 & 0.00 & 0.12 & 0.08 & 0.00 \\
 & \texttt{gemma-2-9b-it} & 0.00 & 0.84 & 0.00 & 0.00 & 0.32 & 0.16 & 0.00 & 0.72 & 0.24 & 0.00 & 0.56 & 0.20 \\
 & \texttt{glm-4-9b-chat-hf} & 0.28 & 0.48 & 0.28 & 0.00 & 0.28 & 0.04 & 0.08 & 0.20 & 0.12 & 0.16 & 0.24 & 0.04 \\
\midrule
\multirow{7}{*}{\textbf{CoT}} & \texttt{Llama-3.1-8B-Instruct} & 0.08 & 0.20 & 0.24 & 0.00 & 0.28 & 0.20 & 0.00 & 0.16 & 0.32 & 0.08 & 0.16 & 0.16 \\
 & \texttt{Mistral-7B-Instruct-v0.3} & 0.20 & 0.04 & 0.04 & 0.00 & 0.00 & 0.20 & 0.08 & 0.00 & 0.08 & 0.04 & 0.00 & 0.12 \\
 & \texttt{Qwen2.5-7B-Instruct} & 0.12 & 0.36 & 0.04 & 0.00 & 0.04 & 0.00 & 0.00 & 0.08 & 0.00 & 0.00 & 0.20 & 0.00 \\
 & \texttt{Yi-1.5-9B-Chat} & 0.56 & 0.36 & 0.16 & 0.04 & 0.32 & 0.12 & 0.24 & 0.12 & 0.04 & 0.20 & 0.36 & 0.04 \\
 & \texttt{deepseek-llm-7b-chat} & 0.44 & 0.08 & 0.12 & 0.04 & 0.04 & 0.00 & 0.16 & 0.04 & 0.00 & 0.12 & 0.00 & 0.04 \\
 & \texttt{gemma-2-9b-it} & 0.48 & 0.44 & 0.04 & 0.00 & 0.20 & 0.04 & 0.00 & 0.56 & 0.04 & 0.12 & 0.52 & 0.00 \\
 & \texttt{glm-4-9b-chat-hf} & 0.16 & 0.20 & 0.00 & 0.00 & 0.12 & 0.00 & 0.00 & 0.04 & 0.00 & 0.04 & 0.00 & 0.00 \\
\midrule
\multirow{7}{*}{\textbf{Enumeration}} & \texttt{Llama-3.1-8B-Instruct} & 0.16 & 0.52 & 0.12 & 0.00 & 0.68 & 0.28 & 0.00 & 0.84 & 0.48 & 0.08 & 0.64 & 0.32 \\
 & \texttt{Mistral-7B-Instruct-v0.3} & 0.24 & 0.04 & 0.08 & 0.00 & 0.12 & 0.08 & 0.08 & 0.16 & 0.00 & 0.00 & 0.08 & 0.08 \\
 & \texttt{Qwen2.5-7B-Instruct} & 0.08 & 0.56 & 0.00 & 0.00 & 0.32 & 0.04 & 0.04 & 0.60 & 0.16 & 0.08 & 0.56 & 0.08 \\
 & \texttt{Yi-1.5-9B-Chat} & 0.12 & 0.80 & 0.24 & 0.04 & 0.84 & 0.08 & 0.16 & 0.84 & 0.20 & 0.16 & 0.60 & 0.16 \\
 & \texttt{deepseek-llm-7b-chat}$^\dagger$ & 0.68 & 0.20 & 0.28 & 0.36 & 0.20 & 0.16 & 0.36 & 0.08 & 0.28 & 0.48 & 0.16 & 0.28 \\
 & \texttt{gemma-2-9b-it} & 0.00 & 1.00 & 0.16 & 0.00 & 0.52 & 0.20 & 0.00 & 0.92 & 0.48 & 0.00 & 0.68 & 0.20 \\
 & \texttt{glm-4-9b-chat-hf} & 0.16 & 0.64 & 0.32 & 0.04 & 0.24 & 0.28 & 0.04 & 0.52 & 0.28 & 0.08 & 0.40 & 0.20 \\
\bottomrule
\end{tabular}
\caption{Breakdown by predicate class (GAP contexts): share of \textit{True} on A (T$_A$), \textit{Neither} on A (N$_A$), and \textit{Neither} on B (N$_B$).}
\label{tab:class}
\end{table*}